\documentclass[5p]{elsarticle}

\usepackage{amsmath,amssymb,amsfonts}
\usepackage{graphicx}
\usepackage{textcomp}

\usepackage{url}
\usepackage{multirow}
\usepackage{makecell}
\usepackage{booktabs}
\usepackage{hyperref}
\usepackage{pifont}
\usepackage{pgfplots}
\pgfplotsset{width=.49\textwidth,compat=1.8}
\tikzset{font=\footnotesize}

\definecolor{pred}{RGB}{255,105,97}
\definecolor{pblue}{RGB}{174,198,207}
\definecolor{pgreen}{RGB}{119,221,119}
\definecolor{pblack}{RGB}{10,10,10}

\newcommand{\cmark}{\ding{51}}

\begin{document}

\begin{frontmatter}

\title{Optimizing the image correction pipeline for pedestrian detection in the thermal-infrared domain}

\author[dse]{Christophe Karam}
\author[lynred]{Jessy Matias}
\author[lynred]{Xavier Breniere}
\author[jocelyn]{Jocelyn Chanussot}

\journal{arXiv}

\affiliation[dse]{
    organization={Data Science Experts},
    city={Grenoble},
    postcode={38000},
    country={France}
}
\affiliation[lynred]{
    organization={Lynred},
    city={Veurey-Voroize},
    postcode={38113},
    country={France}
}
\affiliation[jocelyn]{
    organization={Université Grenoble Alpes, CNRS, Grenoble INP, GIPSA-Lab},
    city={Grenoble},
    postcode={38000},
    country={France}
}

\begin{abstract}
Infrared imagery can help in low-visibility situations such as fog and low-light scenarios, but it is prone to thermal noise and requires further processing and correction. This work studies the effect of different infrared processing pipelines on the performance of a pedestrian detection in an urban environment, similar to autonomous driving scenarios. Detection on infrared images is shown to outperform that on visible images, but the infrared correction pipeline is crucial since the models cannot extract information from raw infrared images. Two thermal correction pipelines are studied, the shutter and the shutterless pipes. Experiments show that some correction algorithms like spatial denoising are detrimental to performance even if they increase visual quality for a human observer. Other algorithms like destriping and, to a lesser extent, temporal denoising, increase computational time, but have some role to play in increasing detection accuracy. As it stands, the optimal trade-off for speed and accuracy is simply to use the shutterless pipe with a tonemapping algorithm only, for autonomous driving applications within varied environments.
\end{abstract}

\end{frontmatter}

\section{Introduction}
\label{sec:intro}
As infrared imagery increasingly integrates the realm of deep learning-based computer vision, especially through applications such as autonomous driving, it becomes crucial to optimize performances and examine shortcomings of existing pipelines. While most passively-cooled infrared sensors geared towards consumer applications produce very noisy raw images, the bulk of the literature deals with pre-corrected images ready for processing, and does not concern itself with any noise reduction steps. Most work tackles improving the detection performance and explores image fusion methods between the infrared and visible domains. In contrast, our work deals with the very first part of the signal processing chain: correcting the raw infrared images to obtain clean images that can be used in object detection. In fact, current infrared sensors are highly sensitive to thermal noise, and often require frequent calibration to remain operational in ideal conditions, which is especially true for passively-cooled microbolometer-type sensors. To counter these thermal noise effects, several post-processing solutions have been developed to correct the resulting images. These pipelines are put in place by sensor manufacturers like Lynred or FLIR, with their main goal being thermal noise correction, also called non-uniformity correction (NUC). In addition to this crucial step, other components in this signal processing chain may involve additional noise reduction, such as stripe noise or bad pixel removal, spatial filtering, and tone-mapping. These pipelines are usually tuned to optimize the visual quality of the output images for a human observer, but in this work, we aim to optimize the image quality from the perspective of a neural-network pedestrian detector. The two perspectives are not necessarily complementary, and algorithms used to make eye-pleasing images may be useless or even detrimental to a neural-network. In this sense, we focus on optimizing the correction pipeline's algorithms and parameters to maximize the performance of a real-time pedestrian detector, in the context of an autonomous driving application. 

We will be using Lynred's infrared sensors and proprietary image correction pipeline, which involves a shutter or shutterless NUC, and several additional noise correction steps. The different pipeline parameters will be tested on a pedestrian detection downstream task, using an in-house infrared-visible urban dataset, collected around Grenoble, France.

\section{Background \& Related Work}
\label{sec:related-work}
\subsection{Infrared Correction Pipelines}

The non-uniformity correction (NUC) is the most important step in infrared image processing, and aims to correct the thermal noise due to the heating of the electronic elements in the sensor. Although some infrared sensors are actively cooled, these are more expensive and geared toward higher-stakes applications (like military ones). For autonomous driving and consumer-level applications, passively cooled are most common. As a result, as the camera is used, it heats up and leads to the build-up of thermal noise. For a constant ambient temperature, two pixels in an image do not have the same response, which constitutes an offset error. On top of that, when the ambient temperature varies, the pixel responses do not vary uniformly, which constitutes a gain error. The correction aims at making the pixel responses uniform with respect to the scene temperature and the ambient temperature, by estimating the error gain and offset. To do so, a calibration step is required. Classically, the manufacturer calibrates their cameras offline using images taken in a controlled environment at set ambient temperatures and scene temperatures. During operation, an online calibration also takes place: the camera shutter closes, acting as a reference uniform-temperature scene \cite{shutter-2014, shutter-ama}. This single reference calibration estimates the offset error, while the gain error is assumed to be linear, and already corrected for during the offline calibration. However, online calibration disrupts the image stream while it takes place, which occurs frequently in order to account for scene variations. As a result, several "shutterless" methods have been developed, such as FLIR's Non-Volatile Flat Field Correction, or Lynred's 3-reference shutterless calibration, in addition to others like \cite{shutterless-2016} or \cite{nuc-shutterless}.

Other components of a correction pipeline can also be considered important, at least for human visual quality, such as stripe noise removal, or destriping \cite{destriping-model, destriping-wavelet}, or the removal of defective pixels \cite{registration-pipe, deadpixels}. More importantly, a component that is both relevant to the human viewing experience and processing by neural networks is tone-mapping. As most raw infrared images are encoded in 16-bits, tone-mapping may be necessary to reduce and adapt the dynamic range of the images to something that can be handled by most screens, and properly treated by pretrained neural networks. From a neural-network perspective, tone-mapping can be seen as a simple input normalization, but the nature of the infrared images makes this a more complicated task. This is similar to the issue that High Dynamic Range (HDR) monitors try to solve in the visible domain, but the dynamics in infrared make this a tougher problem: in RGB, we can map tones to 16 million colors (3x8 bits), while we only have 256 available colors in infrared tone mapping (8 bits). Common tone-mapping algorithms include min-max normalization, standardization (or z-scoring), basic histogram equalization, and Contrast Limited Adaptive Histogram Equalization (CLAHE), among others.

\subsection{Infrared Imaging Applications} \label{sec:lit-infra}

Although developments in computer vision techniques have mainly used images taken by sensors that capture light in the visible range, resulting in grayscale or RGB images, most methods can be transferred to the infrared domain without much loss in efficacy. For example, convolutional object detection models that perform well on RGB images also perform well on infrared images, despite challenges related to low resolution and domain adaption due to the shift in the images' spectral range \cite{cnn-thermal-adaptation}. Notable applications of object detection to infrared images tackles autonomous driving, where detecting objects in the surrounding can be heavily impacted by environmental conditions. For this purpose, some datasets have been curated and made available to the public. The KAIST Multispectral Pedestrian Dataset \cite{kaist}, offers 95k color-thermal pairs and 103,128 total annotations for the pedestrian category, split into person, people, and cyclist. Another dataset was developed by Teledyne FLIR, a manufacturer of thermal imaging infrared cameras, with 15 classes in 26,442 images \cite{flir-adas}.\\
Since the available datasets contain RGB-IR image pairs, most innovative work focuses on methods to fuse information from these two modalities in order to optimize the object detection performance \cite{Wang2022ImprovingRO, piafusion}. Hermann, Ruf, and Beyerer have an interesting take on the IR detection problem, whereby they augment the IR images to more closely resemble RGB images \cite{cnn-thermal-adaptation} for pedestrian detection on KAIST.\\
This work examines the different algorithms of Lynred's established pipeline in order to optimize the algorithms and parameter selection, and does not aim to create novel correction and denoising methods. This thorough examination of an entire infrared processing chain is missing from the literature.

% \subsection{Object Detection}

% Object detection is one of the three main computer vision tasks, along with classification and segmentation. While there are object detection techniques that are not based on deep neural networks, these have largely become outdated and outperformed by more recent breakthroughs. In 2014, the first deep learning-based object detection model was released, R-CNN \cite{rcnn}, followed closely by the first single-shot detector, You Only Look Once (YOLO) \cite{yolov1} in 2015. Since then, the object detection landscape has been largely divided between two-stage detectors like R-CNN, and single-stage detectors like YOLO. Although these models have evolved over time to yield more powerful descendants, like Cascade R-CNN \cite{cascadercnn} and YOLOv4 \cite{yolov4}, these convolutional networks have been the main players in object detection tasks. A notable exception is the recent rise of transformer-based models, namely Swin Transformers \cite{swin} and their extensions. The YOLO and Swin Transformer families constitute the current state-of-the-art.

\section{Methods}
\label{sec:methods}

The main objective of this work is to optimize the infrared image processing pipeline in order to achieve the best detection accuracy with the shortest correction time. The current pipeline, as developed by Lynred, is optimized for obtaining the best visual quality from infrared images. In an object detection setting, the images will not be viewed by humans, but rather processed by a neural network for pedestrian detection. As such, there is no need for visual quality as we understand it, and the pipeline should be re-assessed and re-optimized to produce the best possible real-time object detection results.

\begin{figure*}[ht]
    \centering
    \includegraphics[width=.9\textwidth]{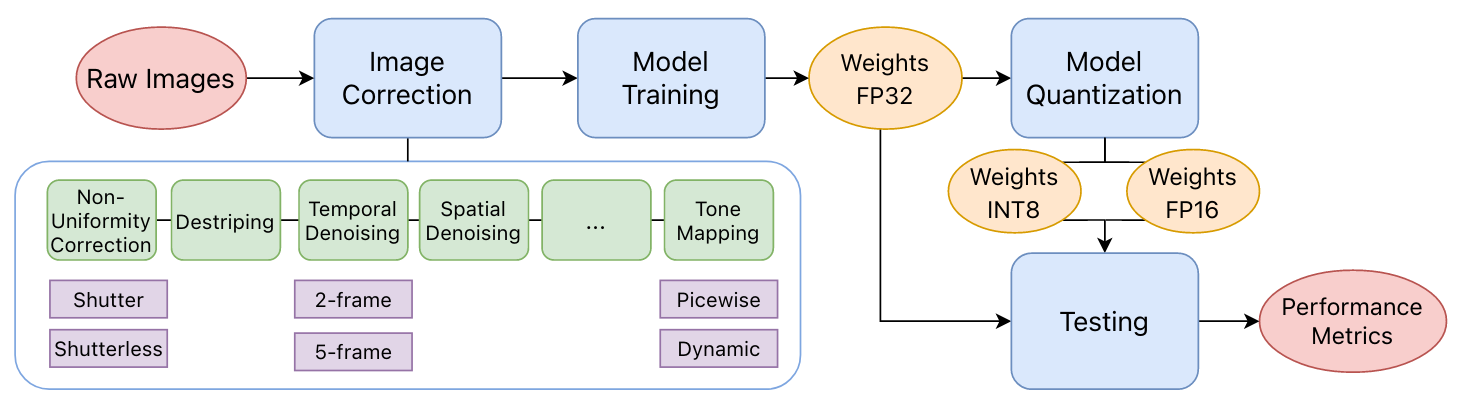}
    \caption{General workflow for optimizing the infrared processing pipeline based on pedestrian detection performance.}
    \label{fig:general_workflow}
\end{figure*}

The workflow adopted in order to achieve this objective contains four main parts, as shown in Figure \ref{fig:general_workflow}. As previously explained, raw infrared images are 16-bit-encoded and very noisy, like in Figure \ref{fig:data_sample} (left). Therefore, the first stage of the workflow is the image correction pipeline, which will ouput images similar to Figure \ref{fig:data_sample} (middle). These corrected images can now serve as an input to train an object detection model, and we get a set of trained weights. These weights are encoded in a 32-bit representation, but can be quantized down to 16-bits, or even 8-bits. This reduces the computational load and memory overhead, thereby speeding up the inference. The final stage of the workflow is to test the different weights on different datasets to evaluate the performance of a given pipeline or setup.

In essence, this work consists in generating datasets from the raw infrared images using different correction pipelines (by tuning algorithm choices and parameters), training a pedestrian detector, then evaluating it on test sets from different datasets for different quantization levels. This large-scale experimentation should allow us to understand the impact of each correction step on the final result in a robust manner.

\vspace{-2mm}

\subsection{Datasets}

\begin{figure}[ht]
    \centering
    \includegraphics[width=.49\textwidth]{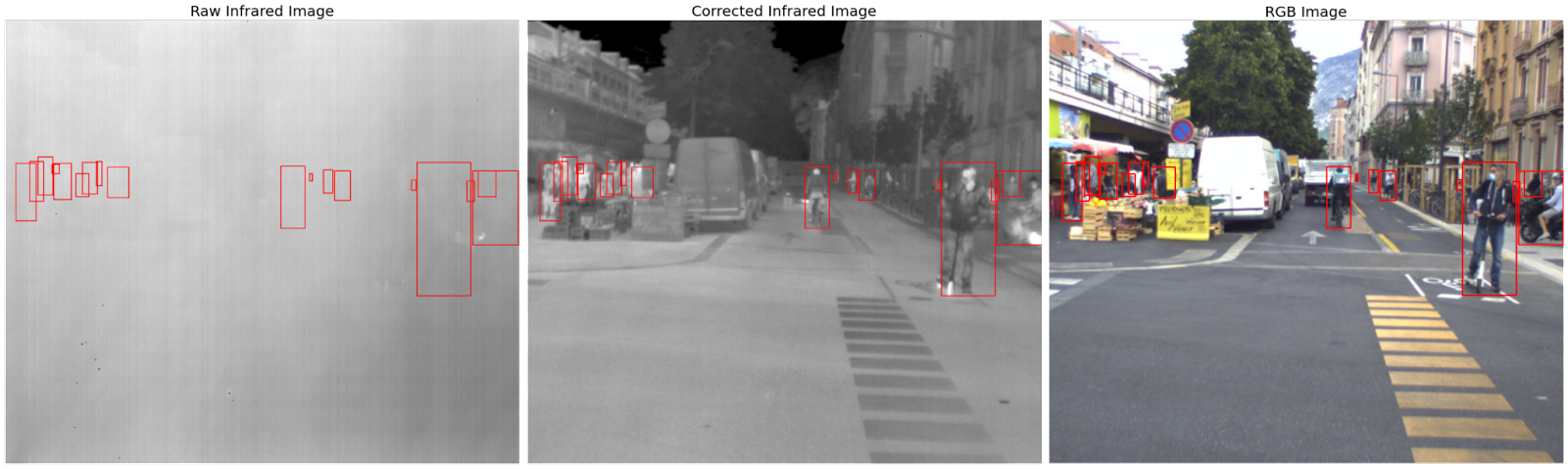}
    \caption{Image with annotated bounding boxes from Lynred VRU dataset across different domains: raw infrared, corrected infrared, and visible RGB.}
    \label{fig:data_sample}
\end{figure}

The main dataset used in this work was provided by Lynred, and called the VRU dataset, for Vulnerable Road Users. It contains 6339 pairs of RGB-infrared images with 24715 annotated bounding boxes for the class "pedestrian" or "person", which includes cyclists and bikers. These infrared images are corrected by Lynred using well-calibrated images, and are 8-bit encoded, like regular images. A subset (5944) of these images is provided in raw 16-bit noisy format. This raw subset will be our main starting point, since we care about applying different correction pipes to these raw images. As can be seen in Figure \ref{fig:data_sample}, the raw images must be corrected to produce usable samples.

Furthermore, two datasets will be used for additional testing, making sure that the conclusions obtained on Lynred's VRU dataset can in fact be generalized to other datasets, making this work useful beyond the scope of a single sensor manufacturer and dataset. These datasets are the Teledyne FLIR ADAS Thermal Dataset v2 \cite{flir-adas}, and finally, the CVC-14 Visible-FIR Day-Night Pedestrian Sequence Dataset \cite{cvc14}. The KAIST Multispectral Pedestrian Detection Benchmark \cite{kaist} dataset was also considered, but its image quality being poor, models trained on it failed to generalize to the other datasets, and vice versa, as will be shown in the upcoming section. As a result, the KAIST dataset was not used for additional testing.

\subsection{Infrared Image Correction}

The entire workflow's goal is to optimize the infrared image correction pipeline, which was developed by Lynred. Lynred provides access to these proprietary correction algorithms in the form of a Software Development ToolKit (SDK). An overview of the correction pipeline is shown in Figure \ref{fig:general_workflow}. As previously stated, the most important part of that pipeline is the non-uniformity correction (NUC), which requires a calibration step to estimate offset and gain errors. For offline calibration, the shutter calibration requires a set of hot and cold images at a constant ambient temperature, and the shutterless pipe requires an additional set of images at another ambient temperature. This 3-reference shutterless calibration allows the correction to become robust to the variation of the sensor temperature during operation. While the NUC step is crucial, additional denoising algorithms are also a main target of this work, since they contribute to the visual quality of the output image, but may be expendable in an object detection workflow. Finally, \textbf{tone-mapping} converts images from uint16 to uint8 representation, for display and processing purposes. In practice, we will first be turning these different correction algorithms on or off to study their impact on object detection. In a second step, we can even vary the parameters and methods used by these algorithms to find the best combination. These algorithms include:

\begin{itemize}
    \item \textbf{Bad Pixel Replacement}: replaces defective image pixels with a weighted mean of their neighborhood, using a 5x5 convolution kernel.
    \item \textbf{Destriping}: removes vertical stripes or column noise present as artifacts on images.
    \item \textbf{Spatial Denoising}: removes spatial noise from image using classical methods such as Recursive Bilateral Filtering \cite{rbf} or Non-Local Means \cite{nlm}, among others.    
    \item \textbf{Temporal Denoising}: removes temporal noise between consecutive frames taking into account the estimation motion for each pixel between these frames.
    \item \textbf{Flare Correction}: removes flare or glare artefacts in images due to noise from light sources.    
\end{itemize}

\subsection{Object Detection Models}

This work mainly uses the YOLOv4 family of object detectors for the downstream task \cite{yolov4}. The Darknet framework is written purely in C and can be compiled on many different architectures \cite{darknet13}, making it easily portable to embedded devices, an important factor for autonomous driving architectures. Furthermore, we heavily rely on the YOLOV4-Tiny model to carry out the different experiments, as it gives us a good trade-off between detection accuracy, inference speed, and training time. Indeed, being able to quickly train models is a crucial factor for this study, and YOLOv4-Tiny allows us to complete an experiment in a little over an hour. To confirm some of our results, we also use Swin Transformer-based object detectors \cite{swin}, which are the current state-of-the-art, but as they are large and slow networks, they are not our main focus. After training, the model weights are converted to Open Neural Network Exchange (ONNX) format, then quantized using TensorRT to fp16 and int8 representations, for faster inference on Nvidia GPUs.

% .

\section{Results}
\label{sec:results}
\subsection{Experimental Setup}

Experiments were performed on a workstation with an Nvidia RTX 3060 GPU (12GB). Training for YOLO models is done through the Darknet framework \cite{darknet13}. Finally, the quantization is handled by TensorRT. Our main evaluation metric is the Average Precision (AP) taken at an Intersection over Union (IoU) threshold at 0.5, called AP@50. Note that all experiments use the YOLOv4-Tiny model trained using the VRU train set, with an input image resolution of $544 \times 448$, unless otherwise specified.

\subsection{Choice of Test Datasets}

\begin{figure}[ht]
    \centering
    \includegraphics[width=.48\textwidth]{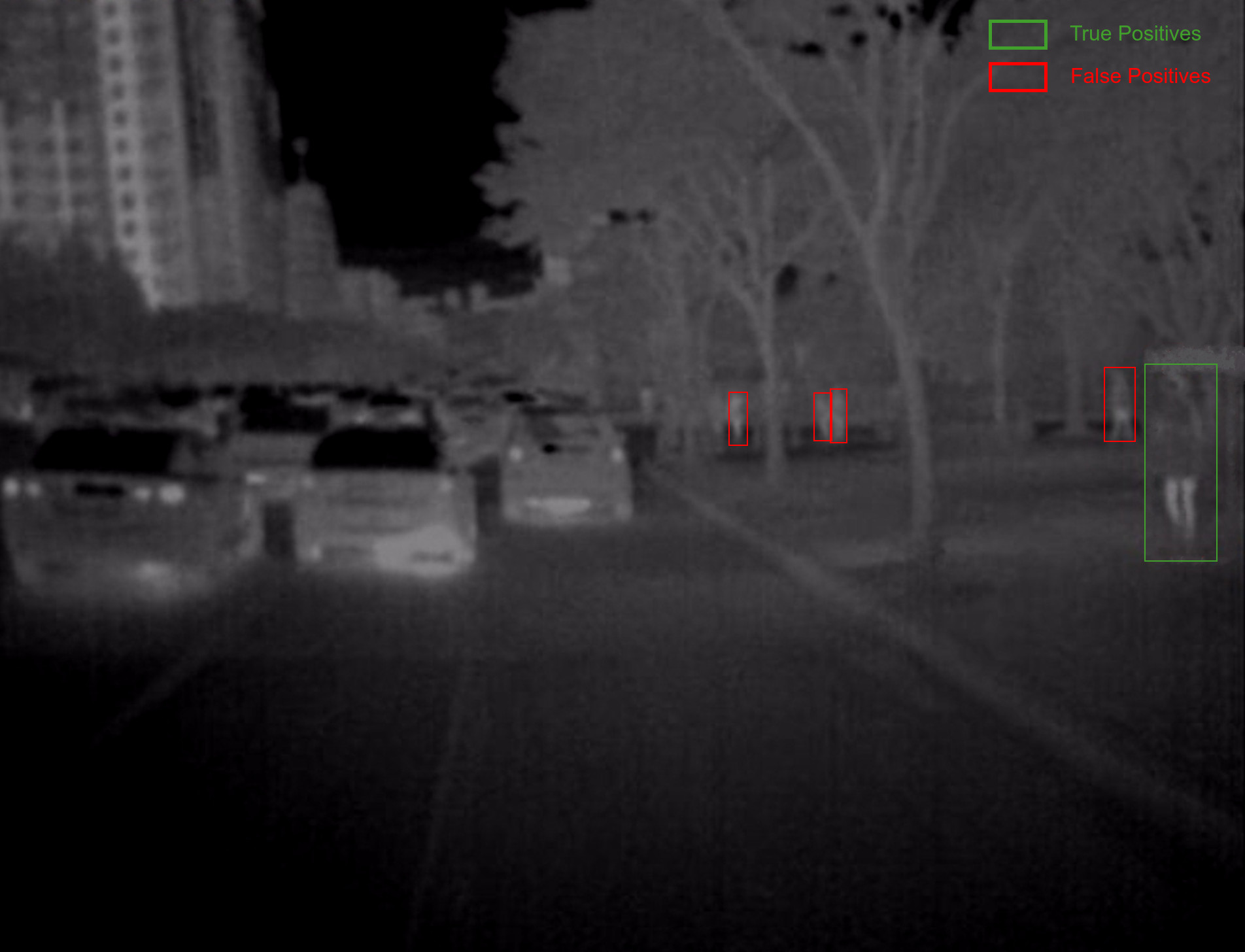}
    \caption{Image from KAIST with our model's predictions, marked as true or false positives based on the actual but erroneous "groundtruth" annotations.}
    \label{fig:kaist_yolo_positives}
\end{figure}

As previously mentioned, our main dataset is Lynred's Vulnerable Road Users (VRU) dataset, while other datasets from the literature include KAIST, FLIR, and CVC. After filtering for pedestrian, FLIR's dataset yields around 21K good quality images, while CVC yields around 6K images, not far from VRU's 6K images as well. On the other hand, KAIST yields around 32K images, but of much lower visual and label quality. To get an idea of the expected performances, a YOLOv4-Tiny model is trained on each of these 4 datasets' training sets, and tested on all of the test sets, with the provided corrected infrared images. The results are shown in Table \ref{tbl:dataset-bench}. We can see that as expected, the values on the diagonal are the highest (mostly), meaning that training and testing on the same dataset produces the best results for that dataset. However, we can also see that models trained on VRU yield the second best performances when testing on other datasets, meaning that models trained on VRU generalize quite well. Furthermore, KAIST seems to interact poorly with the other datasets, due its image and label quality, which is why it won't be included in upcoming experiments. Additionally, it is interesting to note that models trained on VRU achieve higher test metrics on CVC and FLIR than VRU. This difference will persist across all of our experiments, and is due to the difficulty of the VRU dataset. In fact, VRU is mostly collected in the summer, where pedestrians are less contrasted with the background due to the elevated temperatures, and also contains very small objects (e.g. occluded pedestrians in the background), and many targets per image. However, this difficulty comes with variability, making it a good dataset to train robust models able to generalize to different datasets. This will help us confirm some of our hypotheses and results later on. It is important to note that only the VRU dataset changes with every experiment, being regenerated from the raw images through the experiment's pipeline, while the FLIR and CVC datasets do not change and remain the same throughout all experiments. Finally, it may be useful to verify whether the limitation in performance comes from the limited dataset sizes. We combine the quality CVC, FLIR, and VRU train sets and we test the performances of a model trained on this bigger set, as presented in the last row of Table \ref{tbl:dataset-bench}. Only FLIR benefits from this combined dataset, while CVC is mostly unaffected. Performance on VRU drops by 10-points, showing that the task's difficulty is not made easier using a larger dataset within the same domain. As expected, performance on KAIST definitely drops because KAIST is not included in the combined training set, and because its image and label quality are not as good. As an example of the latter, Figure \ref{fig:kaist_yolo_positives} shows our model's prediction on a sample from KAIST. The image quality is worse than the other datasets, and the predictions are wrongly classified as false positives, because the groundtruth annotations are lacking.

\begin{table}[ht]
    \centering
    \scalebox{0.88}{
    \begin{tabular}{cccccc}
        & & \multicolumn{4}{c}{Test Set} \\
        \toprule
        &  & CVC & VRU & FLIR & KAIST \\ 
        \cmidrule(lr){2-6}
        \multirow{5}{*}{\rotatebox[origin=c]{90}{\small Train Set}} & CVC & \textbf{77.17} & 18.17 & 33.46 & 9.17 \\
        & VRU & 65.51 & \textbf{51.99} & 71.38 & 14.04 \\
        & FLIR & 62.70 & 39.18 & 79.15 & 5.04 \\ 
        & KAIST & 18.44 & 7.25 & 32.99 & \textbf{44.64} \\
        & VRU+FLIR+CVC & 76.10 & 41.94 & \textbf{83.29} & 16.9\\  
        \bottomrule
    \end{tabular}}
    \caption{Pedestrian detection AP for cross-training-testing on CVC, VRU, FLIR, and KAIST datasets using YOLOv4-Tiny model}
    \label{tbl:dataset-bench}
\end{table}

\subsection{The role of infrared and correction pipelines}

\begin{table}[ht]
    \centering
    \scalebox{0.6}{
    \begin{tabular}{ccccccccc}
        \Xhline{3\arrayrulewidth}
        Modality & $AP$ & $AP_{day}$ & $AP_{night}$ & $AP_{fp16}$ & $AP_{int8}$ & $t$ (ms) & $t_{fp16}$ (ms) & $t_{int8}$ (ms) \\ 
        \Xhline{3\arrayrulewidth}
        RGB - HR & 51.24 & \textbf{57.57} & 40.58 & 50.76 & 48.76 & 26.92 & 3.70 & 2.58 \\
        RGB & 43.39 & 48.89 & 34.02 & 41.81 & 40.08 & 7.44 & 0.95 & 0.65 \\
        IR & \textbf{51.79} & 38.18 & \textbf{64.45} & 49.53 & 47.82 & 5.69 & 0.96 & 0.65 \\
        IR-Raw & 5.97 & 5.56 & 6.13 & - & - & 5.78 & - & - \\       
        \bottomrule
    \end{tabular}}
    \caption{YOLOv4-Tiny test detection AP@50 on VRU for RGB and IR modalities, with inference times ($t$). All resolutions are at $544 \times 448$ except for RGB-HR which is at $1248 \times 928$}
    \label{tbl:rgb-ir-bench}
\end{table}

The first important point to underline is the usefulness of the infrared modality for pedestrian detection in an urban environment. The model trained on pre-corrected infrared images outperforms the model trained on RGB images at the same medium resolution ($544 \times 448$) by a wide margin of 8.4AP. The IR model even beats a high-resolution RGB model at $1248 \times 928$ by a slight margin, even when the HR images contain more than 4 times more pixels than the medium resolution ones. This is in part explained by the fact that the RGB modality produces a poor representation of pedestrians at night, while infrared is not negatively affected by the time of day. On the contrary, warm-bodied pedestrians will be clearer and contrast even more with the cooler, less busy night-time environment, as evidenced by the day-night metrics in Table \ref{tbl:rgb-ir-bench}. On the other hand, the IR model does not match RGB performance during daytime, in part due to the busy environment leading to images with crowds and numerous objects, and in part due to the hot temperatures during which VRU was collected, making it a tougher dataset because the pedestrians' body temperatures are less contrasted with their environments.

A second major point is the need for a proper correction pipeline. While the raw infrared images look like complete noise to a human observer, the information is still there, and we may hope that a model can disentangle the useful information from the rest of the noise, but that is clearly not the case. Detection on raw infrared images produces extremely poor results, and justifies the need for a correction pipeline to obtain usable images.

We note that the IR-base models will serve as a reference and upper bound for the experiments that will follow. in fact, this dataset is pre-corrected by Lynred using well-calibrated images that match the environmental conditions during which the dataset was collected. Therefore, our offline calibration strategies will not be able to attain or surpass the performances reported here, and our goal will be to get as close as possible, answering the question: how close can a single offline calibration get to online calibration performance?

Finally, an auxiliary solution to the speed vs accuracy tradeoff is model quantization to leverage hardware acceleration, in our case on Nvidia GPUs using TensorRT. We can quantize our YOLO model down to fp16 or int8 representations, which gives us a huge boost in inference time, for the cost of only a few AP points. In Table \ref{tbl:rgb-ir-bench}, we can clearly see that quantizing down generally nets a 7-fold decrease in inference time with a 1-2 point decrease in AP, and that quantizing down nets a 10 or 11-fold decrease in inference time, with a 2-3 point decrease in AP. Since Lynred's infrared processing chain is in 16-bits, then it would make sense to opt for the float 16 quantization as a best compromise between accuracy and speed.

In the following experiments, we will benchmark both the shutter and shutterless pipes to find the optimal set of correction algorithms. It is important to keep in mind that the relative differences in accuracy are the main focus rather than the absolute values.

\subsection{The shutter correction pipeline}

\subsubsection{Calibration Temperatures}

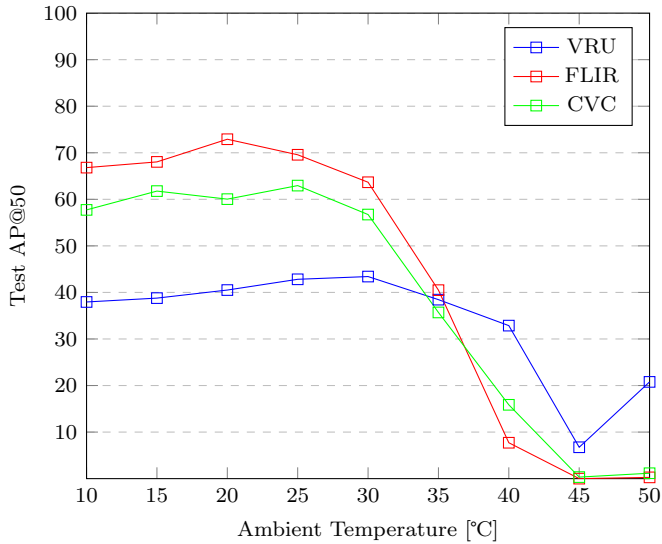
\begin{figure}[h]
\centering
\begin{tikzpicture}
\begin{axis}[
    title={},
    xlabel={Ambient Temperature [\textcelsius]},
    ylabel={Test AP@50},
    xmin=10, xmax=50,
    ymin=0, ymax=100,
    xtick={10,15,20,25,30,35,40,45,50},
    ytick={10,20,30,40,50,60,70,80,90,100},
    legend pos=north east,
    ymajorgrids=true,
    grid style=dashed,
]

\addplot[
    color=blue,
    mark=square,
    ]
    coordinates {
    (10,37.95)(15,38.78)(20,40.51)(25,42.82)(30,43.41)(35,38.45)(40,32.87)(45,6.74)(50,20.78)
    };
\addplot[
    color=red,
    mark=square,
    ]
    coordinates {
    (10,66.81)(15,68.04)(20,72.91)(25,69.57)(30,63.67)(35,40.49)(40,7.71)(45,0.00)(50,0.27)
    };

\addplot[
    color=green,
    mark=square,
    ]
    coordinates {
    (10,57.73)(15,61.77)(20,60.04)(25,62.95)(30,56.73)(35,35.7)(40,15.86)(45,0.33)(50,1.16)
    };
\legend{VRU, FLIR, CVC}        
\end{axis}
\end{tikzpicture}
\caption{Test performances for a varying ambient calibration temperature for a shutter pipe with piecewise tonemapping}
\label{fig:shutter-temps}
\end{figure}

The shutter pipe requires a calibration step using two groups of reference images: one group of images of a scene at $T_{scene-1}$, and another group of images at $T_{scene-2}$, both taken at the same ambient temperature $T_{amb}$. For our experiments, we will always have $T_{scene-1} = 10$ and $T_{scene-2} = 40$, acting as cold and hot references, and ambient temperatures ranging between $10$ and $50$ \textcelsius. In Figure \ref{fig:shutter-temps}, we can see the evolution of the detection performance on our three test sets, VRU, FLIR, and CVC, as a function of the ambient temperature used for calibration. The trends are similar for the three datasets, and provide insight about the variability of environmental conditions during the dataset creation. For FLIR, we see a peak at 20\textcelsius~after which the performance starts to degrade rapidly, dropping by more than 30 points at a temperature of 35\textcelsius. The test on CVC shows the same behavior, but with a peak 25\textcelsius~. On the other hand, tests on the VRU dataset seems more robust to changes in the calibration temperture, with the test metric being relatively stable to within a few percentage points from 10 to 35\textcelsius, peaking at 30\textcelsius. Even after that, performances at high temperatures above 40\textcelsius~are still much better than on the two other datasets. This tells us that the VRU dataset provides more variability in the environmental conditions of its images (wider ranges of temperatures), and that the CVC and FLIR datasets were probably collected in colder weather. We remind that the VRU dataset changes with each experiment, but the CVC and FLIR test sets do not. For the remainder of these experiments, we will be using a calibration ambient temperature of 25\textcelsius, as this would be the average peak temperature across the datasets, keeping these trials fair even if it places us at a slight disadvantage relative to VRU.

\subsubsection{Tonemapping}

\begin{figure}[h]
\centering
\begin{tikzpicture}
\begin{axis}[
    ybar=0.1pt,
    bar width=6pt,
	x tick label style={rotate=45,anchor=east},
	ylabel=Test AP@50,
	enlargelimits=0.1,
    ytick={0,5,10,15,20,25,30,35,40,45,50,55,60,65,70,75},
    ymajorgrids=true,
    grid style=dashed,
	legend style={at={(0.5,1.1)},
	anchor=north,legend columns=-1},
    xtick=data,
    symbolic x coords={piecewise,clip,adaptive1,equalized,adaptive2,clahe,minmax,std3,dynamic},
]
\addplot+[
    style={fill=pblue, draw=none},
    error bars/.cd, 
    y dir=both, 
    y explicit, 
    error bar style={pblack},
    error mark options={
      rotate=90,
      pblack,
      mark size=1.2pt,
      line width=0.5pt
    }]
	coordinates {
        (piecewise,41.99320547540492) +- (0.1,0.35)
        (clip,42.56795698196722) +- (0,0.55)
        (adaptive1,42.686516549163606) +- (0,0.55)
        (equalized,42.692727895268554) +- (0,0.26)
        (adaptive2,42.85312415644722) +- (0,1.02)
        (clahe,42.94335247647692) +- (0,0.69)
        (minmax,43.12371584213261) +- (0,0.22)
        (std3,43.28059582721863) +- (0,0.99)
        (dynamic,43.28567070464053) +- (0,0.32)};
\addplot+[
    style={fill=pred, draw=none},
    error bars/.cd, 
    y dir=both, 
    y explicit, 
    error bar style={pblack},
    error mark options={
      rotate=90,
      pblack,
      mark size=1.2pt,
      line width=0.5pt
    }]
	coordinates {
        (piecewise,69.2235195540357) +- (0,1.12)
        (clip,69.28020854914836) +- (0,1.75)
        (adaptive1,71.02570129623565) +- (0,0.94)
        (equalized,69.48635551545263) +- (0,1.27)
        (adaptive2,69.46849329692273) +- (0,0.49)
        (clahe,69.43058110115659) +- (0,1.50)
        (minmax,70.22864575302314) +- (0,0.31)
        (std3,70.29288194237611) +- (0,0.35)
        (dynamic,70.47938547310649) +- (0,0.90)};
\addplot+[
    style={fill=pgreen, draw=none},
    error bars/.cd, 
    y dir=both, 
    y explicit, 
    error bar style={pblack},
    error mark options={
      rotate=90,
      pblack,
      mark size=1.2pt,
      line width=0.5pt
    }]
	coordinates {
        (piecewise,62.60223387119665) +- =(0,0.39)
        (clip,62.16504160401462) +- (0,0.41)
        (adaptive1,62.21926077281093) +- (0,1.15)
        (equalized,63.442656741989914) +- (0,0.25)
        (adaptive2,65.28505062636708) +- (0,0.69)
        (clahe,61.967011679301834) +- (0,2.55)
        (minmax,63.182878235930104) +- (0,0.61)
        (std3,61.920656060978175) +- (0,1.46)
        (dynamic,61.12051075183673) +- (0,0.69)}; 
\legend{VRU,FLIR,CVC}
\end{axis}
\end{tikzpicture}
\caption{Test performances for a varying tonemapping algorithms with shutter pipe at 25\textcelsius~ambient temperature.}
\label{fig:shutter-tm}
\end{figure}
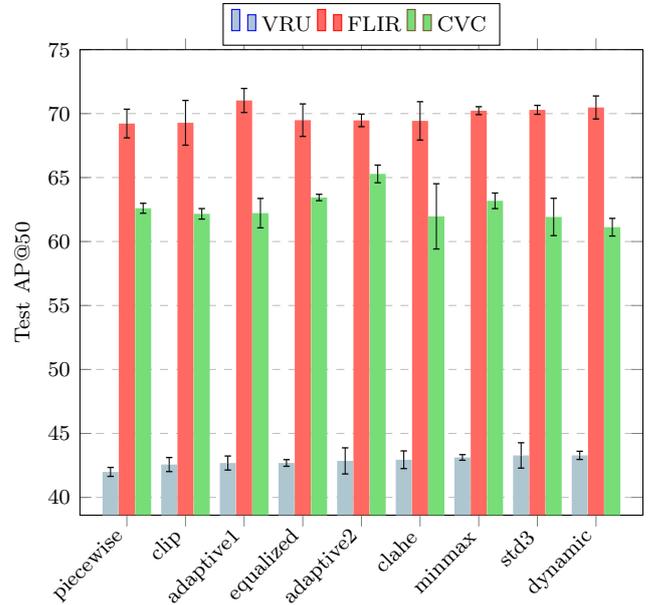

Now that we've determined the best calibration temperatures for our use-case, we can move on to choosing a tonemapping algorithm. It is important to note that virtually all frameworks for training neural networks operate in float32. The raw infrared images in VRU are in uint16 representation, but have a possible dynamic range of $2^{14}$ values. Effectively, when accounting for bad pixels, the real range lies somewhere around $2^{10}$ values. This makes it so that any simple processing such as a naive min-max normalization to convert from uint16 to fp32 cannot succeed consistently on our set of images. After testing basic min-max and z-scoring, we conclude that a basic normalization only produces poor results and that we cannot dispense with tonemapping algorithms carefully conceived to handle infrared imagery. If not, the image histograms end up being squashed and spread along a vast range, diluting the information and rendering the images unusable with our pretrained neural networks. We therefore benchmark the different tonemapping algorithms used by Lynred in their correction pipelines, and the results are shown in Figure \ref{fig:shutter-tm} for the shutter pipe. We can see that for the VRU dataset, the dynamic tonemapping produces the best results, while the piecewise produces the poorest results, with the default settings. On the FLIR dataset, dynamic tonemapping produces the second best results, but it does not perform as well on the CVC dataset. Placing VRU and FLIR as the priorities, as CVC is a much smaller dataset, we choose the dynamic tonemapping for the remainder of these experiments, even if it places us at a slight disadvantage relative to CVC.

\subsubsection{Correction Algorithms}

Finally, having chosen a suitable calibration temperature and tonemapping algorithm, we can test all possible combinations of the different correction algorithms to find the best one. Previously, the shutter pipe only contained the NUC correction step and the tonemapping step. Now, we can include Destriping (Destrip), Bad-Pixel Replacement (BPR), Spatial Denoising (SDN), and Temporal Denoising (TDN). All possible combinations of these 4 algorithms is the power set, leading to 16 total experiments, each run at least 5 times. This includes the empty set, i.e. with only NUC and Tonemapping, which we will use as the baseline with only the two crucial steps. Adding more correction algorithms on top would lead to an increase in the processing chain time, so there is a tradeoff to be tuned: which combinations will allow us to exceed baseline performances, and and what cost in added correction time? These results are presented in Table \ref{tbl:shutter-algs}, with the baseline performances contained between two horizontal lines. The rows are sorted by best VRU-AP, so the rows above the baseline yield better results on VRU, with the best combination highlighted in bold for each test set. However, since we only really care about significant improvements over the baseline, we conduct statistical tests on these results, and highlight in red the ones who are significantly better than the baseline, with a p-value less than 5\%. The first conclusion that can be made is that spatial denoising (SDN) is detrimental to object detection accuracy: in the best-case scenario, whenever Spatial Denoising is included in the correction, it yields a loss of 3\% points for VRU, 13\% points for FLIR, and more than 19\% for CVC, relative to their respective baselines. A second conclusion is that destriping seems to be beneficial: in the 4 combinations where it figures without spatial denoising, and it yields 2 out of 2 significant improvements for VRU, 4 out of 4 significant improvements for FLIR, and 4 out of 7 improvements for CVC. Regarding bad-pixel replacement (BPR), it doesn't seem to provide much value on its own: successful combinations that include it also include destriping, and do not show statistically significant relative differences. This makes sense as a few defective pixels in an image should not present a major obstacle for a well-trained convolutional neural network. Similarly, the benefits of temporal denoising (TDN) are strongly shown in these results. Finally, it is important to take a look at the correction times with these processing chains. The baseline processing with NUC and dynamic tonemapping is executed at an average of 2ms per image. Bad-pixel replacement does not add a significant delay to this time, while temporal denoising adds 1-2ms per image, but requires keeping at least a single previous frame in memory to perform this correction. Destriping, the most useful algorithm, adds around 10-11ms per image, which is quite costly.

% tm dynamic: 2.019 $\pm$ 0.13
% tdn: 3.359 $\pm$ 0.247
% tdn+bpr: 3.38 $\pm$ 0.259
% tdn+destrip: 13.969 $\pm$ 0.502
% tdn+bpr+destrip: 13.951 $\pm$ 0.404
% bpr: 1.984 $\pm$ 0.143
% bpr+destrip: 12.613 $\pm$ 0.383
% destrip: 12.705 $\pm$ 0.687

\begin{table}[t!]
\centering
  \scalebox{0.68}{
  \begin{tabular}{cccc||ccc}
        \Xhline{3\arrayrulewidth}
        Destrip & BPR & SDN & TDN & VRU-AP & FLIR-AP & CVC-AP  \\ 
        \hline
        \cmark & \cmark & & \cmark & \textcolor{red}{\textbf{44.13 $\pm$ 0.51}} & \textcolor{red}{72.34 $\pm$ 0.46} & \textcolor{red}{\textbf{65.26 $\pm$ 0.63}} \\ 
        \cmark & \cmark & & & \textcolor{red}{43.93 $\pm$ 0.39} & \textcolor{red}{\textbf{73.11 $\pm$ 1.45}} & \textcolor{red}{64.04 $\pm$ 0.68} \\
        \cmark & & & \cmark & 43.72 $\pm$ 0.51 & 71.16 $\pm$ 1.41 & \textcolor{red}{64.91 $\pm$ 0.87} \\
        \cmark & & & & 43.27 $\pm$ 0.47 & \textcolor{red}{72.03 $\pm$ 0.95} & \textcolor{red}{63.47 $\pm$ 0.53} \\
        \hline
        & & & & 43.22 $\pm$ 0.62 & 70.06 $\pm$ 0.88 & 61.56 $\pm$ 0.78 \\
        \hline
        & & & \cmark & 43.03 $\pm$ 0.68 & 68.97 $\pm$ 1.00 & \textcolor{red}{63.91 $\pm$ 1.05} \\
        & \cmark & & & 42.11 $\pm$ 1.33 & 70.73 $\pm$ 1.17 & \textcolor{red}{63.03 $\pm$ 1.12} \\
        & \cmark & & \cmark & 42.42 $\pm$ 0.62 & 70.95 $\pm$ 0.91 & \textcolor{red}{63.05 $\pm$ 1.23} \\ 
        \cmark & \cmark & \cmark & & 40.88 $\pm$ 0.61 & 57.70 $\pm$ 3.30 & 42.69 $\pm$ 1.98 \\
        \cmark & \cmark & \cmark & \cmark & 40.80 $\pm$ 0.77 & 54.09 $\pm$ 1.86 & 42.83 $\pm$ 2.39 \\
        \cmark & & \cmark & & 39.97 $\pm$ 0.65 & 54.72 $\pm$ 3.04 & 33.26 $\pm$ 3.25 \\
        & \cmark & \cmark & & 39.89 $\pm$ 0.53 & 56.48 $\pm$ 4.25 & 41.75 $\pm$ 1.89 \\
        & \cmark & \cmark & \cmark & 39.78 $\pm$ 0.29 & 56.91 $\pm$ 5.52 & 42.79 $\pm$ 2.07 \\
        \cmark & & \cmark & \cmark & 39.63 $\pm$ 0.83 & 56.21 $\pm$ 3.67 & 35.89 $\pm$ 2.57 \\
        & & \cmark & & 38.73 $\pm$ 0.64 & 53.91 $\pm$ 1.81 & 32.77 $\pm$ 3.40 \\
        & & \cmark & \cmark & 38.29 $\pm$ 0.64 & 56.97 $\pm$ 1.00 & 35.53 $\pm$ 1.86 \\
        
        \Xhline{3\arrayrulewidth}
    \end{tabular}}
    \label{tbl:shutter-algs}
    \caption{YOLOv4-Tiny detection APs on VRU, FLIR, and CVC for different combinations of correction algorithms for the shutter pipe with Dynamic tone mapping, sorted by best VRU-AP. Values in bold represent the best metrics for a test set, and values between line separators are the baseline performances when no algorithms are added, and values in red are values which are significantly better than the baseline with a p-value less than 5\%.}
\end{table}

\subsection{The shutterless correction pipeline}

\subsubsection{Calibration Temperatures}

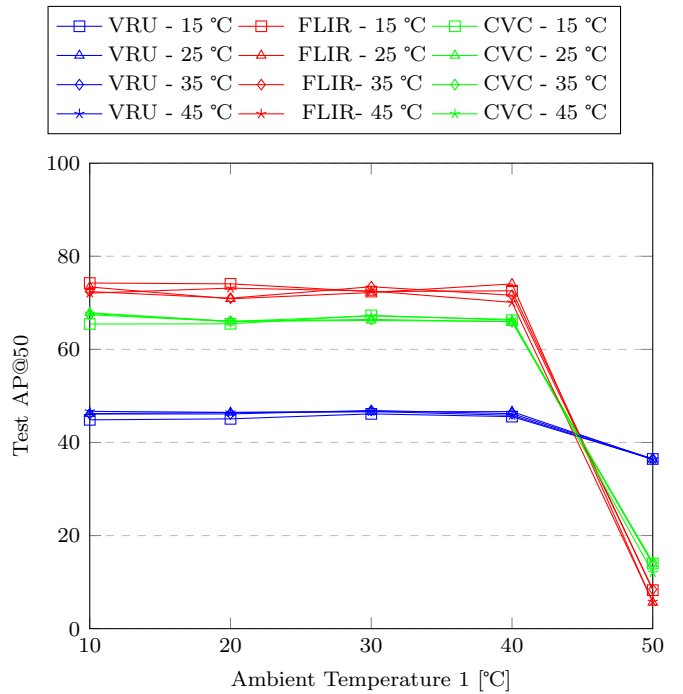
\begin{figure}[h]
\centering
\begin{tikzpicture}
\begin{axis}[
    title={},
    xlabel={Ambient Temperature 1 [\textcelsius]},
    ylabel={Test AP@50},
    xmin=10, xmax=50,
    ymin=0, ymax=100,
    xtick={10,20,30,40,50},
    ytick={0, 20,40,60,80,100},
    legend columns=3,
    legend style={
        /tikz/column 3/.style={
            column sep=3pt,
        },
        at={(-0.075,1.20)},anchor=west
    },
    ymajorgrids=true,
    grid style=dashed,
]

\addplot[color=blue, mark=square] coordinates { (10,44.86)(20,45.07)(30,46.14)(40,45.55)(50,36.45) };
\addlegendentry{VRU - 15 \textcelsius}
\addplot[color=red, mark=square] coordinates { (10,74.25)(20,74.07)(30,72.41)(40,72.58)(50,8.28) };
\addlegendentry{FLIR - 15 \textcelsius}
\addplot[color=green, mark=square] coordinates { (10,65.44)(20,65.49)(30,67.3)(40,66.28)(50,13.99) };
\addlegendentry{CVC - 15 \textcelsius}

\addplot[color=blue, mark=triangle] coordinates { (10,46.19)(20,46.44)(30,46.59)(40,46.63)(50,36.36) };
\addlegendentry{VRU - 25 \textcelsius}
\addplot[color=red, mark=triangle] coordinates { (10,73.4)(20,70.86)(30,72.2)(40,74.06)(50,5.63) };
\addlegendentry{FLIR - 25 \textcelsius}
\addplot[color=green, mark=triangle] coordinates { (10,67.86)(20,66.03)(30,66.2)(40,65.99)(50,13.91) };
\addlegendentry{CVC - 25 \textcelsius}

\addplot[color=blue, mark=diamond] coordinates { (10,46.11)(20,46.13)(30,46.87)(40,46.19)(50,36.49) };
\addlegendentry{VRU - 35 \textcelsius}
\addplot[color=red, mark=diamond] coordinates { (10,72.41)(20,70.99)(30,73.45)(40,71.63)(50,8.45) };
\addlegendentry{FLIR- 35 \textcelsius}
\addplot[color=green, mark=diamond] coordinates { (10,67.44)(20,66.0)(30,66.44)(40,65.93)(50,14.23) };
\addlegendentry{CVC - 35 \textcelsius}

\addplot[color=blue, mark=star] coordinates { (10,46.68)(20,46.46)(30,46.67)(40,45.83)(50,36.36) };
\addlegendentry{VRU - 45 \textcelsius}
\addplot[color=red, mark=star] coordinates { (10,72.08)(20,73.16)(30,72.53)(40,70.1)(50,5.79) };
\addlegendentry{FLIR- 45 \textcelsius}
\addplot[color=green, mark=star] coordinates { (10,67.69)(20,65.99)(30,67.17)(40,66.43)(50,12.01) };
\addlegendentry{CVC - 45 \textcelsius}

\end{axis}
\end{tikzpicture}
\caption{Varying ambient temperatures for a shutterless calibration with piecewise tonemapping, trained on VRU, with a cold scene temperature of 10\textcelsius ~for ambient temperature 2 (in legend).}
\label{fig:shutterless-temps}
\end{figure}

The shutterless pipe takes a little longer to correct images than the shutter pipe, but produces higher quality images for a single factory calibration. In fact, online shutter calibration is costly in terms of frame rate, but frequently calibrates the camera to its true operating conditions. A shutterless pipe allows us to improve the image quality without having to resort to online calibration. Just like the shutter pipe, shutterless calibration needs two groups of images at temperatures ($T_{amb-1}$, $T_{scene-1}$), ($T_{amb-1}$, $T_{scene-2}$), but also requires an additional set of images at another reference ambient temperature, ($T_{amb-2}$, $T_{scene-3}$). In our case, the scene temperatures options are only $T_{scene-1} = 10$ and $T_{scene-2} = 40$\textcelsius, so $T_{scene-3}$ will also either be $10$ or $40$\textcelsius. The ambient temperatures can vary between $10$ and $50$\textcelsius. All in all, we will have three possible temperatures to vary: $T_{amb-1}$, $T_{amb-2}$, and $T_{scene-3}$. Figure \ref{fig:shutterless-temps} shows test metrics on VRU, FLIR, and CVC for $T_{amb-1} = \{10, 20, 30, 40, 50\}$ , $T_{amb-2} = \{15,25,35,45\}$, and $T_{scene-3} = 10$\textcelsius. The same configuration with $T_{scene-3} = 40$\textcelsius yields very similar results, so it won't be shown here to keep the graph simple. We can clearly see that the trends we saw earlier with the shutter pipe temperatures in Figure \ref{fig:shutter-temps} are now flattened out. The shutterless pipe has rendered the performance less sensitive to the initial temperature calibration, and allows for a robust detection in varying conditions. The only significant performance drop that remains is at 50\textcelsius, which is to be expected for such an extreme temperature. The choice of calibration temperature is no longer as important, but the best compromise between all datasets seems to be for $T_{amb-1} = 30$\textcelsius , $T_{amb-2} = 35$\textcelsius, and $T_{scene-3} = 10$\textcelsius.

% \begin{table}
%     \centering
%     \scalebox{0.82}{
%     \begin{tabular}{ccccccc}
%          & & \multicolumn{5}{c}{} \\
%          \toprule
%          & & $T_{a1} = 10$ & $T_{a1} = 20$ & $T_{a1} = 30$ & $T_{a1} = 40$ & $T_{a1} = 50$ \\
%          \cmidrule(lr){1-7}
%          \multirow{4}{*}{\rotatebox[origin=c]{90}{\small $T_{sc} = 10$}} & $T_{a2} = 15$ & XX & XX &  XX & XX & XX\\
%          & $T_{a2} = 25$ & XX & XX & XX & XX & XX\\
%          & $T_{a2} = 35$ & XX & XX & XX & XX & XX\\
%          & $T_{a2} = 45$ & XX & XX & XX & XX & XX\\
%         \cmidrule(lr){1-7}
%         \multirow{4}{*}{\rotatebox[origin=c]{90}{\small $T_{sh} = 40$}} & $T_{a2} = 15$ & XX & XX &  XX & XX & XX\\
%         & $T_{a2} = 25$ & XX & XX & XX & XX & XX \\
%         & $T_{a2} = 35$ & XX & XX & XX & XX & XX\\
%         & $T_{a2} = 45$ & XX & XX & XX & XX & XX\\
%         \bottomrule
%     \end{tabular}}
%     \label{tbl:shutterless-temp-bench}
%     % \caption{Benchmark for shutterless calibration image temperatures using YOLOv4-Tiny on VRU Test, with three references: \\$T_{a1} = T_{ambient-1}$ at $T_{sc} = T_{scene-cold} = 10$ \\ and \\ $T_{a1} = T_{ambient-1}$ at $T_{sh} = T_{scene-hot} = 40$ \\ and \\ $T_{a2} = T_{ambient-2}$ \\ at \\ $T_{sc} = T_{scene-cold} = 10$ OR $T_{sc} = T_{scene-hot} = 40$}    
%     \caption{Benchmark for shutterless calibration image temperatures using YOLOv4-Tiny on VRU Test, with three references, with $T_{a1} = T_{ambient-1}$, $T_{a2} = T_{ambient-2}$, $T_{sh} = T_{scene-hot}$, $T_{sc} = T_{scene-col}$}
% \end{table}

\subsubsection{Tonemapping}

Just as for the shutter pipe, after we've determined our calibration temperatures, we can move on to choosing a tonemapping algorithm. Unlike before, dynamic tonemapping is no longer a top-performer for VRU, so we will take the std3 algorithm instead. Another good choice across all datasets and both pipes would have been the minmax algorithm. In fact, both methods are similar: the minmax algorithm performs a min-max normalization with user-set minimum and maximum values, whereas the std3 algorithm also performs the same normalization but with the minimum and maximum values set to be $mean_{pixels} \pm 3\cdot std_{pixels}$. For most, the test performance on VRU isn't wildly affected, but the differences in FLIR and CVC metrics are a bit more pronounced. We will proceed with std3 for the rest of these shutterless experiments.

\begin{figure}[h]
\centering
\begin{tikzpicture}
\begin{axis}[
    ybar=0.1pt,
    bar width=6pt,
	x tick label style={rotate=45,anchor=east},
	ylabel=Test AP@50,
	enlargelimits=0.1,
    ytick={0, 10, 20, 30, 40, 50, 60, 70},
    ymajorgrids=true,
    grid style=dashed,
	legend style={at={(0.5,1.1)},
	anchor=north,legend columns=-1},
    xtick=data,
    symbolic x coords={adaptive2,dynamic,piecewise,adaptive1,clahe,equalized,minmax,clip,std3},
]
\addplot+[
    style={fill=pblue, draw=none},
    error bars/.cd, 
    y dir=both, 
    y explicit, 
    error bar style={pblack},
    error mark options={
      rotate=90,
      pblack,
      mark size=1.2pt,
      line width=0.5pt
    }]
	coordinates {
        (adaptive2,45.13513793651131) +- (0,1.00)
        (dynamic,45.31036390025159) +- (0,1.30)
        (piecewise,46.59157173262816) +- (0,0.23)
        (adaptive1,46.6235177747875) +- (0,0.43)
        (clahe,46.80528442030858) +- (0,0.27)
        (equalized,46.86159221278419) +- (0,0.46)
        (minmax,46.88570510340869) +- (0,0.18)
        (clip,46.90019542029868) +- (0,0.78)
        (std3,46.97508020641454) +- (0,0.20)};
\addplot+[
    style={fill=pred, draw=none},
    error bars/.cd, 
    y dir=both, 
    y explicit, 
    error bar style={pblack},
    error mark options={
      rotate=90,
      pblack,
      mark size=1.2pt,
      line width=0.5pt
    }]
	coordinates {
        (adaptive2,71.04522325068244) +- (0,0.86)
        (dynamic,72.34659013727834) +- (0,0.60)
        (piecewise,72.30087920889588) +- (0,0.50)
        (adaptive1,71.86785514955415) +- (0,1.04)
        (clahe,72.57107329721273) +- (0,1.51)
        (equalized,72.38694380159151) +- (0,1.04)
        (minmax,72.94026242209148) +- (0,0.50)
        (clip,72.3921749826673) +- (0,1.53)
        (std3,72.04815046195485) +- (0,0.64)};
\addplot+[
    style={fill=pgreen, draw=none},
    error bars/.cd, 
    y dir=both, 
    y explicit, 
    error bar style={pblack},
    error mark options={
      rotate=90,
      pblack,
      mark size=1.2pt,
      line width=0.5pt
    }]
	coordinates {
        (adaptive2,67.60457100097547) +- (0,0.58)
        (dynamic,67.28467605010131) +- (0,0.60)
        (piecewise,66.79551235797133) +- (0,0.54)
        (adaptive1,67.85018550972201) +- (0,0.44)
        (clahe,67.02086541073315) +- (0,0.86)
        (equalized,66.85693178859667) +- (0,0.95)
        (minmax,68.44481484177925) +- (0,1.07)
        (clip,66.25945944521794) +- (0,0.43)
        (std3,66.11550899498599) +- (0,0.09)}; 
\legend{VRU,FLIR,CVC}
\end{axis}
\end{tikzpicture}
\label{fig:shutterless-tm}
\caption{Test performances for a varying tonemapping algorithms with a shutterless pipe at 30-35-10\textcelsius~calibration temperatures.}
\end{figure}
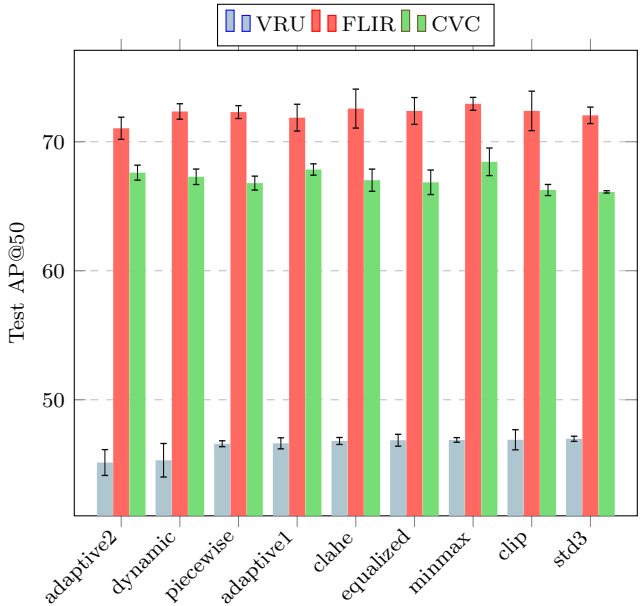

\subsubsection{Correction Algorithms}

\begin{table}[t!]
\centering
  \scalebox{0.62}{
  \begin{tabular}{ccccc||ccc}
        \Xhline{3\arrayrulewidth}
        Destrip & BPR & SDN & TDN & Flare & VRU-AP & FLIR-AP & CVC-AP  \\ 
        \hline      
        \cmark & \cmark &  &  &  & \textbf{47.54 $\pm$ 0.74} & \textbf{73.43 $\pm$ 1.23} & 66.46 $\pm$ 1.86 \\
        \cmark &  &  & \cmark &  & \textcolor{red}{47.12 $\pm$ 0.25} & 72.98 $\pm$ 0.68 & \textcolor{red}{67.39 $\pm$ 0.71} \\
        \cmark & \cmark &  & \cmark &  & 47.07 $\pm$ 0.59 & 73.40 $\pm$ 0.52 & 65.31 $\pm$ 2.04 \\
        \cmark & \cmark &  &  & \cmark & 46.90 $\pm$ 0.39 & 72.71 $\pm$ 0.64 & 67.01 $\pm$ 1.08 \\
        &  &  & \cmark &  & 46.81 $\pm$ 0.39 & 72.43 $\pm$ 0.92 & 66.87 $\pm$ 1.07 \\
        \hline
        &  &  &  &  & 46.72 $\pm$ 0.29 & 73.02 $\pm$ 1.25 & 66.23 $\pm$ 1.24 \\
        \hline
        \cmark &  &  & \cmark & \cmark & 46.67 $\pm$ 1.26 & 73.27 $\pm$ 1.18 & 67.52 $\pm$ 2.11 \\
        & \cmark &  &  &  & 46.59 $\pm$ 0.60 & 72.50 $\pm$ 1.34 & 65.58 $\pm$ 1.41 \\
        \cmark &  &  &  & \cmark & 46.58 $\pm$ 0.50 & 72.66 $\pm$ 1.70 & 67.71 $\pm$ 1.52 \\
         & \cmark &  & \cmark &  & 46.57 $\pm$ 0.58 & 72.56 $\pm$ 0.97 & 65.62 $\pm$ 1.95 \\
         \cmark &  &  &  &  & 46.31 $\pm$ 0.41 & 73.03 $\pm$ 0.91 & 66.45 $\pm$ 1.13 \\
        \cmark & \cmark &  & \cmark & \cmark & 46.28 $\pm$ 0.92 & 71.97 $\pm$ 0.51 & 66.76 $\pm$ 2.27 \\
         &  &  & \cmark & \cmark & 46.27 $\pm$ 0.48 & 72.13 $\pm$ 1.01 & 67.08 $\pm$ 0.84 \\
         &  &  &  & \cmark & 45.97 $\pm$ 0.65 & 71.49 $\pm$ 1.38 & 67.18 $\pm$ 0.90 \\
         & \cmark &  & \cmark & \cmark & 45.95 $\pm$ 0.93 & 72.33 $\pm$ 0.87 & 67.03 $\pm$ 1.98 \\
         & \cmark &  &  & \cmark & 45.57 $\pm$ 0.60 & 70.86 $\pm$ 1.17 & \textcolor{red}{\textbf{67.95 $\pm$ 0.46}} \\
        \cmark & \cmark & \cmark &  &  & 44.06 $\pm$ 0.85 & 62.89 $\pm$ 0.48 & 44.33 $\pm$ 1.84 \\
        \cmark & \cmark & \cmark & \cmark &  & 44.00 $\pm$ 0.72 & 59.27 $\pm$ 2.92 & 43.28 $\pm$ 2.79 \\
        \cmark &  & \cmark & \cmark &  & 43.95 $\pm$ 0.31 & 63.16 $\pm$ 1.90 & 39.27 $\pm$ 2.03 \\
        \cmark &  & \cmark &  &  & 43.81 $\pm$ 0.54 & 62.63 $\pm$ 1.62 & 39.34 $\pm$ 3.30 \\
         & \cmark & \cmark &  &  & 43.75 $\pm$ 1.04 & 64.57 $\pm$ 1.21 & 43.30 $\pm$ 1.13 \\
         & \cmark & \cmark & \cmark &  & 43.58 $\pm$ 0.22 & 64.40 $\pm$ 2.23 & 46.69 $\pm$ 1.42 \\
         & \cmark & \cmark & \cmark & \cmark & 43.23 $\pm$ 0.47 & 63.03 $\pm$ 1.22 & 48.42 $\pm$ 1.87 \\
         &  & \cmark & \cmark &  & 43.22 $\pm$ 0.44 & 62.70 $\pm$ 2.46 & 38.61 $\pm$ 0.88 \\
        \cmark & \cmark & \cmark &  & \cmark & 43.21 $\pm$ 0.55 & 59.61 $\pm$ 3.26 & 44.88 $\pm$ 1.06 \\
        &  & \cmark &  &  & 43.05 $\pm$ 0.51 & 61.90 $\pm$ 2.88 & 39.67 $\pm$ 2.60 \\
        \cmark & \cmark & \cmark & \cmark & \cmark & 42.92 $\pm$ 0.21 & 60.50 $\pm$ 3.04 & 44.74 $\pm$ 1.52 \\
         & \cmark & \cmark &  & \cmark & 42.78 $\pm$ 0.66 & 63.04 $\pm$ 1.39 & 46.24 $\pm$ 1.49 \\
        \cmark &  & \cmark & \cmark & \cmark & 42.77 $\pm$ 0.48 & 58.45 $\pm$ 2.14 & 34.71 $\pm$ 1.03 \\
        \cmark &  & \cmark &  & \cmark & 42.76 $\pm$ 0.77 & 62.45 $\pm$ 2.32 & 35.72 $\pm$ 3.65 \\
         &  & \cmark &  & \cmark & 42.23 $\pm$ 0.69 & 62.53 $\pm$ 2.66 & 38.05 $\pm$ 2.24 \\
         &  & \cmark & \cmark & \cmark & 41.61 $\pm$ 0.55 & 64.50 $\pm$ 1.15 & 43.06 $\pm$ 2.28 \\
        \Xhline{3\arrayrulewidth}
    \end{tabular}}
    \caption{YOLOv4-Tiny detection APs on VRU, FLIR, and CVC for different combinations of correction algorithms for the shutterless pipe with std3 tonemapping, sorted by best VRU-AP. Values in bold represent the best metrics for a test set, and values between line separators are the baseline performances when no algorithms are added, and values in red are values which are significantly better than the baseline with a p-value less than 5\%.}
    \label{tbl:shutterless-algs}
\end{table}

With our chosen calibration temperatures and tonemapping algorithm, we move on to benchmarking the different correction algorithms presented earlier, namely: Bad Pixel Replacement (BPR), Destriping (Destrip), Flare Correction (Flare), Spatial Denoising (SDN), Temporal Denoising (TDN). The impact of including each of these algorithms in the processing chain or not is shown in Table \ref{tbl:shutterless-algs}. Including spatial denoising seems to have a negative impact on the results, while the impact of bad-pixel replacement seems to be rather neutral. Indeed, bad-pixel replacement is a punctual pixel-level correction that should not lead to differences in the downstream network, while spatial denoising results in a more blurry image, leading to a better visual quality for humans, but a loss in accuracy on small objects that can no longer be distinguished from the backgrounds. Conversely, destriping removes column noise that might be a nuisance to small objects, while flare correction can reduce noise from strong light sources like the sun, leading to a more contrasted image with pedestrians more visible against the background. Unlike the shutter results, we have a lot less significant improvements over the baselines here. The main reason is that as we improve the baseline image quality with a better offline NUC, the other correction algorithms no longer have a major role to play. In terms of correction times, the shutterless baseline is at around 32ms per image, with destriping adding 10-11ms, flare 8ms, temporal denoising 1-2ms, and negligible additions for bad-pixel replacement. For the shutter pipe, it might have been worth it to include destriping even with a 10ms penalty per image, but for the shutterless case, even if combinations including destriping yield the best results across datasets, the differences are not large enough to warrant the added correction time, especially considering that the shutterless pipe is already slower. So, the full shutterless pipe with all included algorithms required a correction time of 57ms per image, and by stripping out all but the NUC and tonemapping, we get back down to 32ms. We also gain in accuracy, going from 42.92 AP with all algorithms included to 46.72 AP with the slimmed-down version.

\section{Conclusion}
\label{sec:Conclusion}
This work examined the intricacies of Lynred's raw infrared image correction pipeline in order to optimize its runtime and subsequent object detection performance. The use of infrared images in autonomous driving applications, especially for pedestrian detection, is highly warranted, as infrared detection outperforms detection in the visible domain. The use of a correction pipeline is also crucial due to the inability of current models to extract information from noisy raw infrared images. When interrupting the video stream is not possible due to safety reasons, such as in the automotive market, the shutterless thermal correction outperforms the factory shutter correction. Within these two pipes, tone-mapping stands out as crucial. Spatial denoising seem to degrade performances, while remaining algorithms, namely Flare, Bad-Pixel Removal and Temporal Denoising do not seem to improve detection and slow down the correction, even if they do in fact improve visual quality for a human observer. Destriping seems to slightly improve detection especially in the shutter pipe, where its impact on correction time can be absorbed by the quick calibration step. In the slower shutterless pipe, destriping does not yield improvements significant enough to warrant its use. The best trade-off in that case is then to opt for a lean processing pipe with NUC and tone-mapping for the shutterless calibration, yielding a 44\% improvement in correction speed and 8\% increase in detection AP over the full pipeline with all algorithms on the VRU dataset. For the shutter pipe, by only including destriping, we shave 4ms off the correction time,  gaining 1AP, 2AP, and 3AP for the VRU, FLIR, and CVC test sets, respectively. On top of that, quantizing the model down to float-16 representation nets a 7-fold increase in inference time, further increasing the responsiveness in real-time applications.

More experimentation can help confirm whether some algorithms truly provide benefits, but it would also be beneficial to examine improving the performance of some algorithms, like tone-mapping. Finally, we could look at the possibilities of using data augmentation to improve the robustness of the shutter pipe to temperature variations, which would result in a substantial speed improvement over the shutterless pipe.

\section{Acknowledgments}
This work was supported by the Fondation Grenoble INP, DeepRed chair, under the patronage of Lynred, and also partially supported by MIAI @ Grenoble Alpes (ANR-19-P3IA-0003)

\bibliographystyle{ieeetr}
\bibliography{refs.bib}

\begin{thebibliography}{10}

\bibitem{shutter-2014}
P.~W. Nugent, J.~A. Shaw, and N.~J. Pust, ``{Radiometric calibration of infrared imagers using an internal shutter as an equivalent external blackbody},'' {\em Optical Engineering}, vol.~53, no.~12, p.~123106, 2014.

\bibitem{shutter-ama}
H.~Budzier and G.~Gerlach, ``1.1 - calibration of infrared cameras with microbolometers,'' pp.~889--894, 01 2015.

\bibitem{shutterless-2016}
A.~Tempelhahn, H.~Budzier, V.~Krause, and G.~Gerlach, ``Shutter-less calibration of uncooled infrared cameras,'' {\em Journal of Sensors and Sensor Systems}, vol.~5, pp.~9--16, 01 2016.

\bibitem{nuc-shutterless}
C.~Liu~et. al., ``Shutterless non-uniformity correction for long-term stability of uncooled long-wave infrared camera,'' {\em Measurement Science and Technology}, vol.~29, 11 2017.

\bibitem{destriping-model}
M.~Li~et. al., ``A novel stripe noise removal model for infrared images,'' {\em Sensors}, vol.~22, p.~2971, 04 2022.

\bibitem{destriping-wavelet}
M.~Li~et. al., ``An infrared stripe noise removal method based on multi-scale wavelet transform and multinomial sparse representation,'' {\em Computational Intelligence and Neuroscience}, vol.~2022, 05 2022.

\bibitem{registration-pipe}
E.~Lielāmurs, A.~Cvetkovs, R.~Novickis, and K.~Ozols, ``Infrared image pre-processing and ir/rgb registration with fpga implementation,'' {\em Electronics}, vol.~12, no.~4, 2023.

\bibitem{deadpixels}
C.~T. Nguyen, N.~Mould, and J.~L. Regens, ``Dead pixel correction techniques for dual-band infrared imagery,'' {\em Infrared Physics \& Technology}, vol.~71, pp.~227--235, 2015.

\bibitem{cnn-thermal-adaptation}
C.~Herrmann, M.~Ruf, and J.~Beyerer, ``{CNN-based thermal infrared person detection by domain adaptation},'' in {\em Autonomous Systems: Sensors, Vehicles, Security, and the Internet of Everything} (M.~C. Dudzik and J.~C. Ricklin, eds.), vol.~10643, pp.~38 -- 43, International Society for Optics and Photonics, SPIE, 2018.

\bibitem{kaist}
S.~H. et. al., ``Multispectral pedestrian detection: Benchmark dataset and baselines,'' in {\em Proceedings of IEEE Conference on Computer Vision and Pattern Recognition (CVPR)}, 2015.

\bibitem{flir-adas}
TELEDYNE, ``Teledyne flir free adas thermal dataset v2,'' 01 2019.
\newblock [Online; posted 19-January-2019].

\bibitem{Wang2022ImprovingRO}
W.~et. al., ``Improving rgb-infrared object detection by reducing cross-modality redundancy,'' {\em Remote. Sens.}, vol.~14, p.~2020, 2022.

\bibitem{piafusion}
T.~et. al., ``Piafusion: A progressive infrared and visible image fusion network based on illumination aware,'' {\em Information Fusion}, vol.~83-84, pp.~79--92, 2022.

\bibitem{cvc14}
A.~Gonzalez~Alzate, Z.~Fang, Y.~Socarras, J.~Serrat, D.~Vázquez, J.~Xu, and A.~López, ``Pedestrian detection at day/night time with visible and fir cameras: A comparison,'' {\em Sensors}, vol.~16, p.~820, 06 2016.

\bibitem{rbf}
Q.~Yang, ``Recursive bilateral filtering,'' vol.~7572, pp.~399--413, 10 2012.

\bibitem{nlm}
A.~Buades, B.~Coll, and J.-M. Morel, ``A non-local algorithm for image denoising,'' in {\em 2005 IEEE Computer Society Conference on Computer Vision and Pattern Recognition (CVPR'05)}, vol.~2, pp.~60--65 vol. 2, 2005.

\bibitem{yolov4}
A.~Bochkovskiy, C.~Wang, and H.~M. Liao, ``Yolov4: Optimal speed and accuracy of object detection,'' {\em CoRR}, vol.~abs/2004.10934, 2020.

\bibitem{darknet13}
J.~Redmon, ``Darknet: Open source neural networks in c.'' \url{http://pjreddie.com/darknet/}, 2013.

\bibitem{swin}
Z.~L. et. al., ``Swin transformer: Hierarchical vision transformer using shifted windows,'' {\em CoRR}, vol.~abs/2103.14030, 2021.

\end{thebibliography}

\end{document}